\documentclass[conference]{IEEEtran}
\IEEEoverridecommandlockouts
\usepackage{cite}
\usepackage{amsmath,amssymb,amsfonts}
\usepackage{algorithmic}
\usepackage{graphicx}
\usepackage{textcomp}
\usepackage{xcolor}
\usepackage[a4paper, total={184mm,239mm}]{geometry}
\def\BibTeX{{\rm B\kern-.05em{\sc i\kern-.025em b}\kern-.08em
\kern-.1667em\lower.7ex\hbox{E}\kern-.125emX}}

\usepackage{censor}
\StopCensoring

\begin{document}

\title{TMIQ: Quantifying Test and Measurement Domain Intelligence in Large Language Models\\

\thanks{\blackout{Emmanuel Olowe is sponsored by Keysight Technologies, United States. Computing resources were partly supported by Royal Society grant RGS\textbackslash R2\textbackslash 222007.}}
}

 \author{\IEEEauthorblockN{\blackout{Emmanuel A. Olowe}}
\IEEEauthorblockA{\textit{\blackout{School of Engineering}} \\
\textit{\blackout{The University of Edinburgh}}\\
\blackout{Edinburgh, UK} \\
\blackout{e.a.olowe@sms.ed.ac.uk}}
\and
\IEEEauthorblockN{\blackout{Danial Chitnis}}
\IEEEauthorblockA{\textit{\blackout{School of Engineering}} \\
\textit{\blackout{The University of Edinburgh}}\\
\blackout{Edinburgh, UK} \\
\blackout{d.chitnis@ed.ac.uk}}
}

\maketitle

\begin{abstract}
The Test and Measurement domain, known for its strict requirements for accuracy and efficiency, is increasingly adopting Generative AI technologies to enhance the performance of data analysis, automation, and decision-making processes. Among these, Large Language Models (LLMs) show significant promise for advancing automation and precision in testing. However, the evaluation of LLMs in this specialized area remains insufficiently explored. To address this gap, we introduce the Test and Measurement Intelligence Quotient (TMIQ), a benchmark designed to quantitatively assess LLMs across a wide range of electronic engineering tasks. TMIQ offers a comprehensive set of scenarios and metrics for detailed evaluation, including SCPI command matching accuracy, ranked response evaluation, Chain-of-Thought Reasoning (CoT), and the impact of output formatting variations required by LLMs on performance. In testing various LLMs, our findings indicate varying levels of proficiency, with exact SCPI command match accuracy ranging from around 56\% to 73\%, and ranked matching first-position scores achieving around 33\% for the best-performing model. We also assess token usage, cost-efficiency, and response times, identifying trade-offs between accuracy and operational efficiency.  Additionally, we present a command-line interface (CLI) tool that enables users to generate datasets using the same methodology, allowing for tailored assessments of LLMs. TMIQ and the CLI tool provide a rigorous, reproducible means of evaluating LLMs for production environments, facilitating continuous monitoring and identifying strengths and areas for improvement, and driving innovation in their selections  for applications within the Test and Measurement industry.
\end{abstract}

\begin{IEEEkeywords}
Artificial intelligence, Large language models, Synthetic LLM Benchmark, Test and measurement, Electronic engineering, Automation
\end{IEEEkeywords}

\footnote{The full benchmark is available: \blackout{https://github.com/labiium/tmiq}}
\section{Introduction}

The Test and Measurement industry is fundamental to electronic engineering, ensuring that instruments and systems meet stringent standards of precision and efficiency. As electronic systems become increasingly complex, there is a growing demand for intelligent automation to maintain high levels of performance and reliability. Large Language Models (LLMs), such as GPT-3 \cite{Brown2020LanguageLearners} and GPT-4 \cite{OpenAI2023GPT-4Report}, have shown remarkable abilities in understanding and generating human-like text. Trained on vast datasets, these models can perform tasks ranging from natural language understanding to complex reasoning and problem-solving. A key advancement enhancing their reasoning capabilities is Chain-of-Thought (CoT) prompting \cite{Wei2022Chain-of-ThoughtModels}, which allows models to generate intermediate reasoning steps. CoT improves performance on tasks requiring multi-step reasoning by enabling the model to break down problems into sequential steps.
Despite these advancements, applying LLMs to specialized domains like Test and Measurement remains underexplored. Tasks such as generating precise Standard Commands for Programmable Instruments (SCPI) or reasoning within constrained engineering contexts demand high accuracy and domain-specific knowledge. Errors in command execution or reasoning can lead to significant consequences, highlighting the need for reliable LLM performance in these critical applications.
Existing evaluation frameworks for LLMs primarily focus on general-purpose tasks and lack the specificity required for specialized domains \cite{Alzahrani2024WhenLeaderboards}. Benchmarks like the Massive Multitask Language Understanding (MMLU) \cite{Hendrycks2020MeasuringUnderstanding}, and its successor \cite{Wang2024MMLU-Pro:Benchmark} have been instrumental in assessing broad language abilities. Model families such as Gemini, Claude and GPT-4 have wide interest in their general performance \cite{Kevian2024CapabilitiesUltra} but there is less exploration of results in domains requiring specialized knowledge such as electronic engineering.

To address this gap, we introduce the Test and Measurement Intelligence Quotient (TMIQ), a benchmark specifically designed to evaluate LLM performance on domain-specific electronic engineering knowledge and SCPI command accuracy. TMIQ assesses an LLM's ability to generate accurate SCPI commands, perform structured reasoning using CoT, and handle output formatting variations impacting usability. We also provide a command-line interface (CLI) tool for users to generate custom datasets, enhancing the benchmark's flexibility and reproducibility. By aligning closely with the demands of the Test and Measurement industry, TMIQ aims to foster the development of AI-driven solutions by facilitating LLM selection in light of for stringent requirements for and operational efficiency.

\section{Methodology}
% Description of your proposed method

% \subsection{EEMT Data Creation Pipeline}
% \begin{figure}[!t]
%     \centering
%     \includegraphics[width=0.5\textwidth]{figures/tmiq.png}
%     \caption{EEMT Benchmark creation process}
%     \label{fig:creation_pipeline}
% \end{figure}

The EEMT (Electronics-Engineering Multiple-choice test) datasets were the series of electronic engineering datasets that were created to test domain-specific knowledge of LLMs. These were synthetic datasets created using LLMs and additional context of domain-specific knowledge

\subsubsection{Mining PDF Data}
The initial phase of the TMIQ benchmark involves converting electronic engineering-related PDFs into Markdown format. This conversion process is critical as it forms the foundation for subsequent data mining using LLMs. PDFs files pertaining to Electronic Engineering and Test and Measurement were obtained using Marker\cite{githubGitHubVikParuchurimarker}.  Marker handles OCR and text extraction tools that preserve the structure and integrity of the original content as much as possible; this includes header and footer removal.

\subsubsection{Generating Question-Answer Pairs}
Once the relevant sections have been extracted, LLMs are employed to generate initial question-answer pairs. The goal is to create a diverse set of questions that thoroughly test the LLMs' understanding of the content across various complexity levels. These questions are formulated to cover both fundamental and advanced concepts within the subdomains, ensuring a comprehensive assessment of the LLMs' capabilities. The generated answers are then carefully reviewed to ensure they accurately reflect the correct interpretation of the content, providing a reliable baseline for further evaluation. A poor answer in this stage is tolerable as a ranking-based methodology is used for the performance metric.

\subsubsection{Mining False Answers}
To challenge the LLMs further, the next step involves generating a set of incorrect answers, or distractors, for each question. These distractors are crafted to be subtly incorrect, with varying degrees of deviation from the correct answer. The incorrect answers are designed to increase in their degree of error, ranging from slight inaccuracies to completely erroneous interpretations. This gradient of incorrectness is intended to assess the LLMs' ability to differentiate between closely related concepts and to test their precision in reasoning and decision-making. The generation of these distractors is also automated using LLMs, which are instructed to produce plausible but incorrect alternatives based on the context of the original content. 
The steps for generating questions-answer pairs and mining false answers were kept separate because combing them into a single step generally lead to increased failure rates in the generating the required JSON output. LLMs would make more mistakes in the JSON being outputted which represents the Question-Answers-False that were required. Additionally, it was found that when combing the tasks, the generated questions tended to be simpler-resulting in only a basic question-answer pair being produced. The number of False Answers targeted for the LLMs to generated for this dataset were 20 but LLMs being non-deterministic could generate more or less.

\subsection{Test Metrics}
The final stage of the TMIQ benchmark involves evaluating the LLMs based on their ability to rank the answers based on their correctness. The LLMs would only have to rank an answer by giving the index based on where it appeared in order (e.g. "Answer: 1, 5, 2, 4, 3"). Two performance metrics were used to grade the LLMs: First Mtch (FM), which awards full marks if the most correct answer is ranked first, and Position Match (PM), which assigns an exponentially decreasing score the further the correct answer is from the first position. These metrics collectively are intended to provide a good assessment of the LLMs' suitability for deployment in real-world test and measurement scenarios within electronic engineering.

To further contextualize model performance, we propose an efficiency metric that combines accuracy, cost, and execution time. 
Accuracy \( S \) is normalized as \( S_{\text{norm}} = \frac{S}{100} \), while the computational cost \( C \), the total API cost in dollars, is calculated based on the number of input tokens \( T_{\text{in}} \), output tokens \( T_{\text{out}} \), and their respective rates for tokens consumed and produced \( r_{\text{in}} \) and \( r_{\text{out}} \):
\[
C = (T_{\text{in}} \times r_{\text{in}}) + (T_{\text{out}} \times r_{\text{out}})
\]
The cost is then normalized as:
\[
C_{\text{norm}} = \frac{C_{\text{max}} - C}{C_{\text{max}} - C_{\text{min}}}
\]
Similarly, execution time \( T \) is normalized as:
\[
T_{\text{norm}} = \frac{T_{\text{max}} - T}{T_{\text{max}} - T_{\text{min}}}
\]
The overall efficiency \( E \) is the geometric mean of the normalized metrics:
\[
E = \left( S_{\text{norm}} \times C_{\text{norm}} \times T_{\text{norm}} \right)^{1/3}
\]
This metric ensures that accuracy, cost, and time contribute equally, penalizing models that perform poorly in any dimension.

\subsection{SCPI Commands Question Pipeline}
\begin{figure*}[!t]
    \centering
    \includegraphics[width=0.8\textwidth]{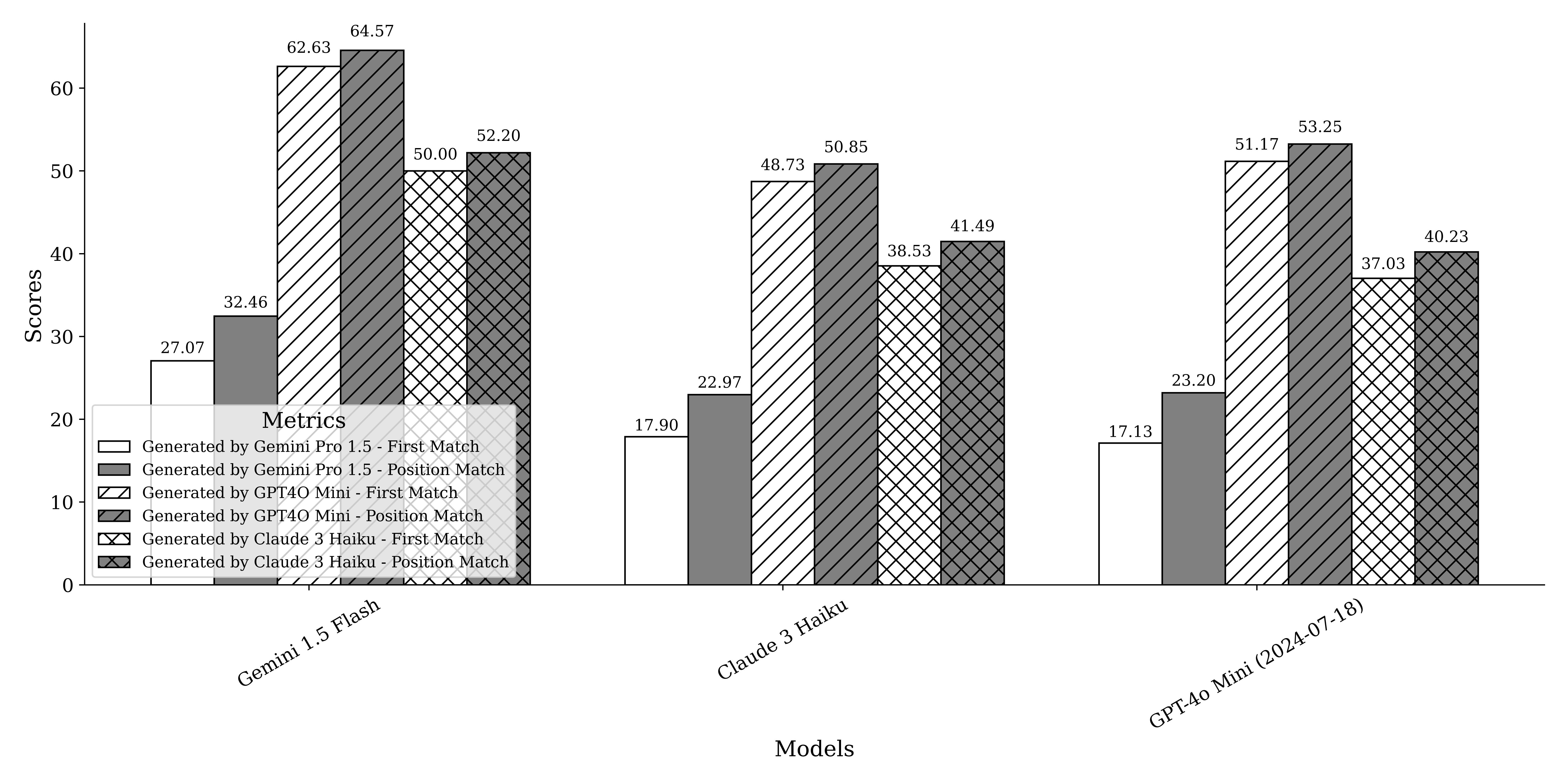}
    \caption{Comparison of difficulty and results of synthetic generation data compared from different models}
    \label{fig:family}
\end{figure*}

\begin{figure*}[!t]
    \centering
    \includegraphics[width=0.8\textwidth]{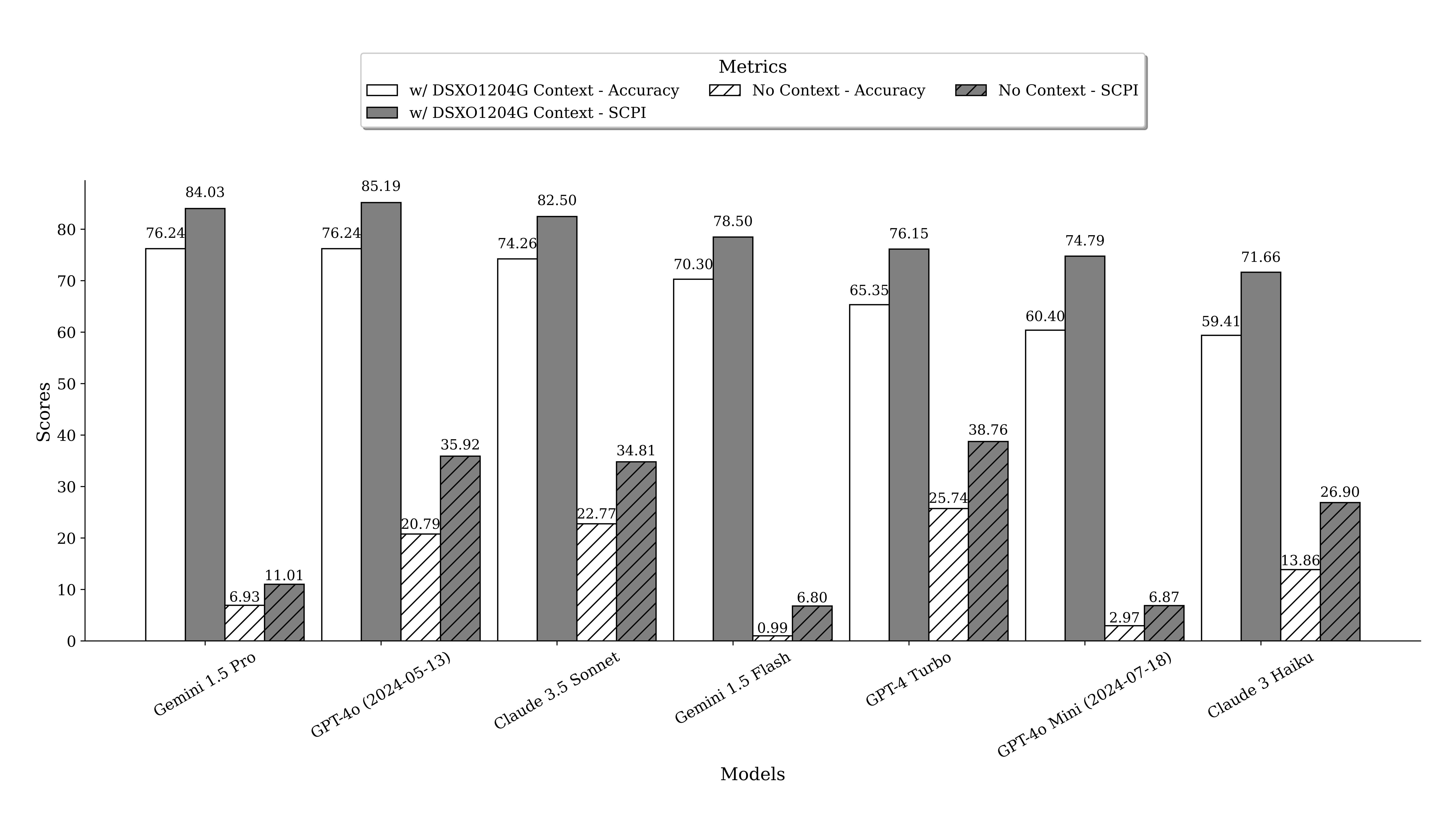}
    \caption{SCPI Evaluation Test with DSOX1204G SCPI Programming Manual Context and without context}
    \label{fig:scpi_correct}
\end{figure*}

\begin{figure*}[!t]
    \centering
    \includegraphics[width=0.8\textwidth]{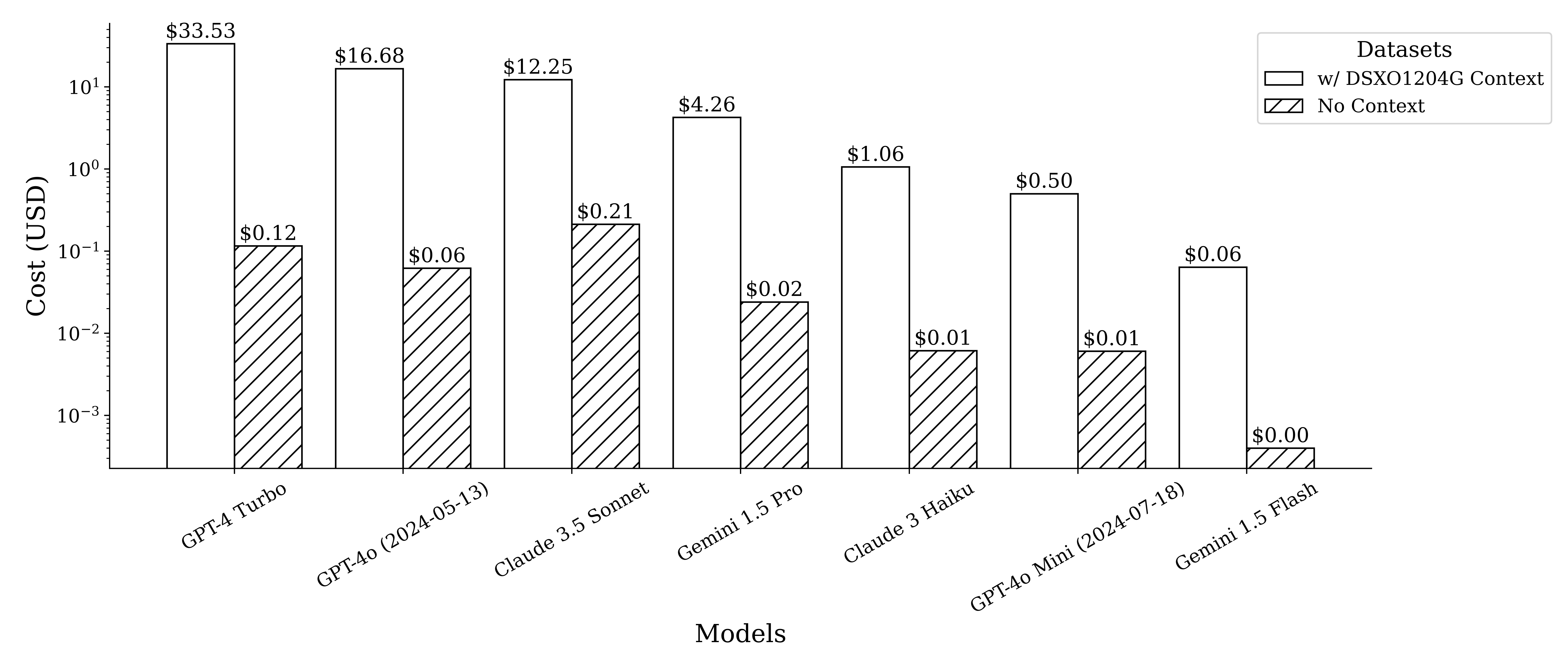}
    \caption{Total cost to complete the SCPI Evaluation Tasks}
    \label{fig:scpi_cost}
\end{figure*}

\begin{figure*}[!t]
    \centering
    \includegraphics[width=0.8\textwidth]{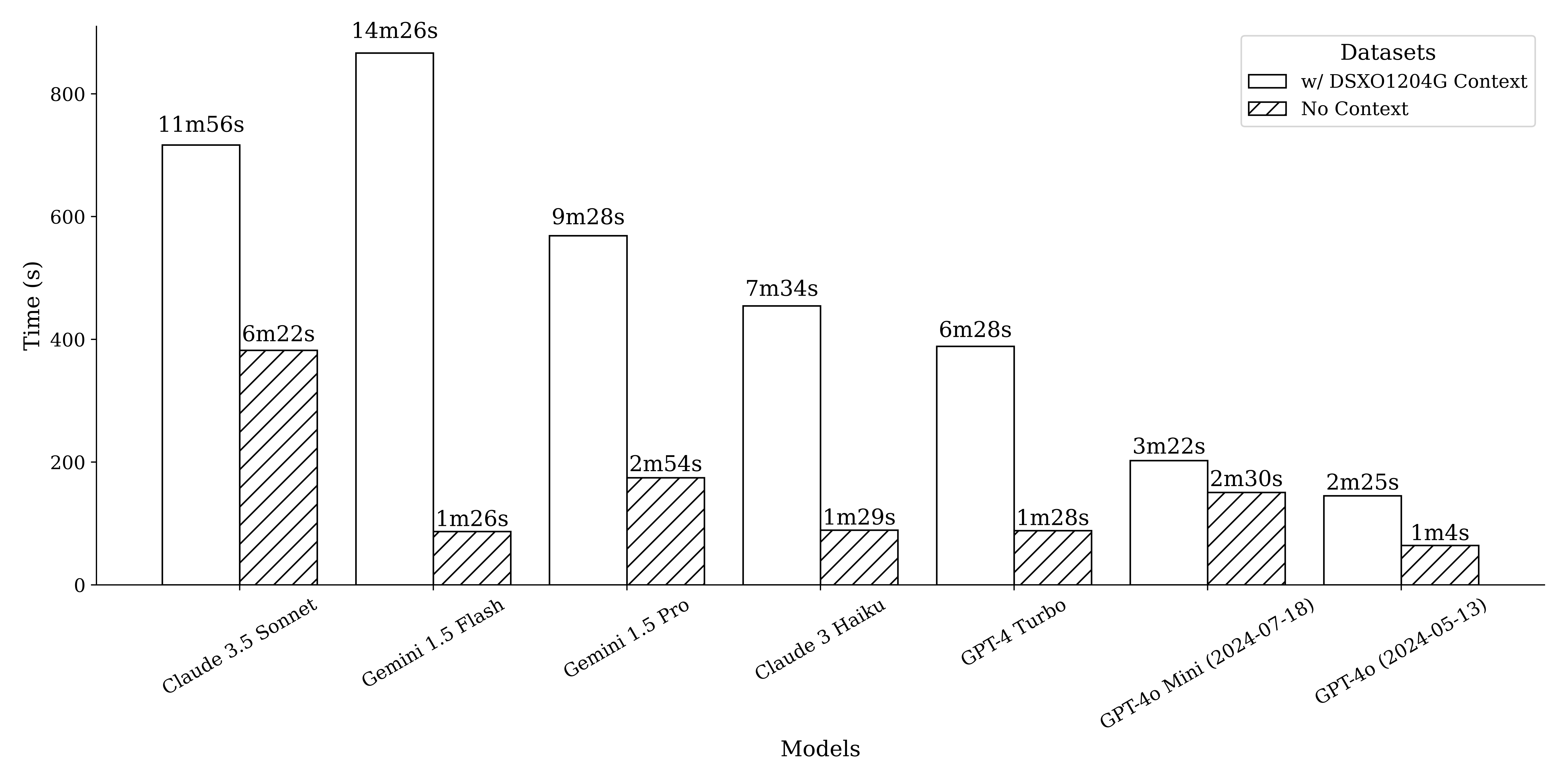}
    \caption{Total LLM Inference time to complete the SCPI Evaluation Tasks}
    \label{fig:scpi_time}
\end{figure*}

\begin{figure*}[!t]
    \centering
    \includegraphics[width=0.8\textwidth]{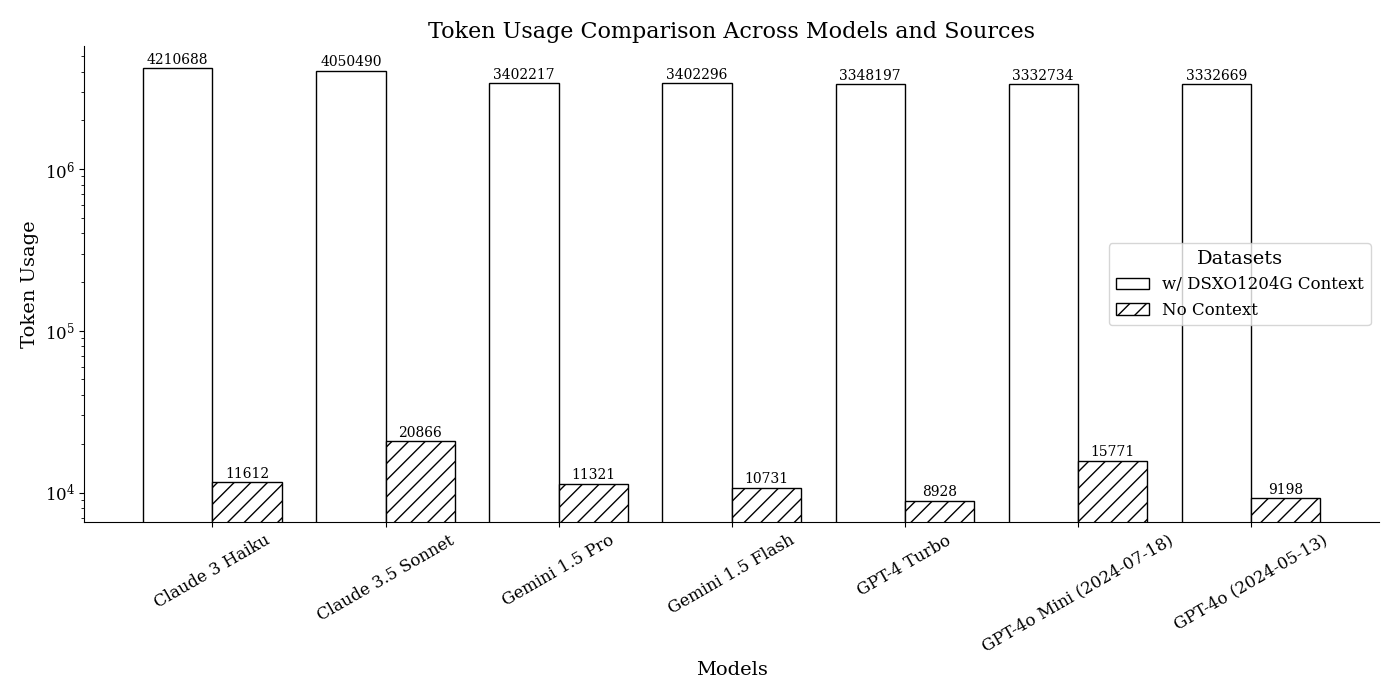}
    \caption{Total Token used to complete the SCPI Evaluation Tasks}
    \label{fig:scpi_tokens}
\end{figure*}

\begin{table*}[ht]
\centering
\scriptsize
\label{Table1}
\begin{tabular}{|l|c|c|c|c|c|c|c|c|c|c|c|c|}
\hline
\textbf{Tag} & \multicolumn{2}{c|}{\textbf{Claude 3 Haiku}} & \multicolumn{2}{c|}{\textbf{Claude 3.5 Sonnet}} & \multicolumn{2}{c|}{\textbf{GPT-4o (05-13)}} & \multicolumn{2}{c|}{\textbf{GPT-4o Mini (07-18)}} & \multicolumn{2}{c|}{\textbf{Gemini 1.5 Flash}} & \multicolumn{2}{c|}{\textbf{Gemini 1.5 Pro}} \\
\hline & \textbf{CoT} & \textbf{no CoT} & \textbf{CoT} & \textbf{no CoT} & \textbf{CoT} & \textbf{no CoT} & \textbf{CoT} & \textbf{no CoT} & \textbf{CoT} & \textbf{no CoT} & \textbf{CoT} & \textbf{no CoT} \\
\hline
\textbf{Analog}\cite{fernandez2017analog} & 9.58 & 19.76 & \textbf{38.92} & \textbf{34.13} & 22.75 & 23.95 & 15.57 & 20.36 & 17.96 & 26.35 & 2.99 & 27.54 \\
\hline
\textbf{DigitalSignal}\cite{smith1999scientist} & 4.17 & 11.90 & \textbf{26.79} & \textbf{28.57} & 16.67 & 16.07 & 10.12 & 10.71 & 11.31 & 21.43 & 0.60 & 16.07 \\
\hline
\textbf{Embedded}\cite{lee2017introduction} & 5.99 & 14.97 & \textbf{31.74} & \textbf{32.93} & 16.77 & 20.96 & 14.97 & 16.17 & 12.57 & 22.75 & 0.00 & 24.55 \\
\hline
\textbf{Instrumentation}\cite{gregory1981introduction} & 3.59 & 11.98 & \textbf{29.34} & \textbf{32.34} & 20.36 & 16.17 & 7.19 & 14.37 & 11.98 & 23.95 & 2.40 & 16.77 \\
\hline
\textbf{InstrumentationFundamentals}\cite{placko2007fundamentals} & 5.39 & 13.77 & \textbf{30.54} & \textbf{30.54} & 18.56 & 19.16 & 13.77 & 16.77 & 14.97 & 27.54 & 0.60 & 23.95 \\
\hline
\textbf{InstrumentationMeasurement}\cite{northrop2005introduction} & 2.99 & 14.37 & \textbf{22.75} & 21.56 & 14.37 & 14.97 & 7.19 & 11.38 & 11.98 & \textbf{22.16} & 0.60 & 19.76 \\
\hline
\textbf{InstrumentationPrinciples}\cite{morris2001measurement} & 2.40 & 11.98 & \textbf{20.96} & \textbf{22.16} & 12.57 & 17.96 & 7.78 & 9.58 & 10.18 & 17.37 & 1.20 & 19.16 \\
\hline
\textbf{IoT}\cite{hanes2017iot} & 6.59 & 30.54 & \textbf{41.32} & \textbf{37.72} & 30.54 & 28.74 & 16.17 & 23.35 & 26.95 & 32.93 & 5.39 & 25.15 \\
\hline
\textbf{OpAmp}\cite{coughlin2001operational} & 6.59 & 20.36 & \textbf{34.13} & \textbf{37.72} & 21.56 & 30.54 & 16.17 & 25.15 & 17.37 & 34.73 & 2.40 & 31.74 \\
\hline
\textbf{Parametric Measurements}\cite{keysight2020parametric} & 2.61 & 15.65 & \textbf{30.43} & \textbf{35.65} & 17.39 & 16.52 & 12.17 & 13.91 & 16.52 & 19.13 & 1.74 & 24.35 \\
\hline
\textbf{RF}\cite{razavi1998rf} & 8.28 & 18.62 & \textbf{31.03} & \textbf{34.48} & 17.24 & 21.38 & 13.10 & 15.86 & 17.24 & 28.28 & 0.69 & 23.45 \\
\hline
\textbf{RFCircuitDesign}\cite{davis2001radio} & 8.98 & 19.76 & \textbf{36.53} & \textbf{41.92} & 17.96 & 23.35 & 14.97 & 20.36 & 19.16 & 28.74 & 1.20 & 29.94 \\
\hline
\textbf{RFCircuitDesign2}\cite{bowick2008rf} & 8.38 & 22.16 & \textbf{40.12} & \textbf{39.52} & 24.55 & 29.94 & 16.17 & 20.36 & 20.36 & 32.34 & 0.00 & 29.94 \\
\hline
\textbf{RFComponents}{\cite{carr2002rf}} & 5.37 & 14.77 & \textbf{31.54} & \textbf{33.56} & 20.81 & 21.48 & 12.08 & 11.41 & 14.77 & 22.15 & 0.00 & 24.83 \\
\hline
\textbf{Semiconductor}{\cite{fiore2021semiconductor}} & 7.19 & 24.55 & \textbf{44.31} & \textbf{39.52} & 27.54 & 31.14 & 20.96 & 28.14 & 24.55 & 36.53 & 2.40 & 32.34 \\
\hline
\textbf{SignalIntegrity}{\cite{resso2009signal}} & 5.39 & 20.96 & \textbf{38.92} & \textbf{38.32} & 20.36 & 27.54 & 13.77 & 18.56 & 15.57 & 28.14 & 0.60 & 25.15 \\
\hline
\textbf{VLSI}{\cite{kang2003cmos}} & 4.19 & 11.38 & \textbf{31.74} & \textbf{34.13} & 17.37 & 19.16 & 8.98 & 12.57 & 10.18 & 28.14 & 1.80 & 27.54 \\
\hline
\textbf{EEML}{\cite{simeone2018machine}} & 7.62 & 21.90 & \textbf{31.43} & \textbf{30.48} & 17.14 & 20.95 & 7.62 & 18.10 & 13.33 & 27.62 & 0.00 & 23.81 \\
\hline
\textbf{renewable}\cite{quaschning2005renewable} & 4.08 & 21.77 & \textbf{35.37} & 31.29 & 21.77  & 21.77 & 10.20 & 17.01 & 17.01 & \textbf{31.97} & 0.68 & 25.85 \\
\hline
\textbf{Avg FM Score} & 5.77 & 17.90 & \textbf{33.13} & \textbf{33.53} & 19.90 & 22.33 & 12.70 & 17.13 & 16.03 & 27.07 & 1.37 & 24.87 \\
\hline
\textbf{Total Errors} & 10231 & 138 & 1 & 3014 & 2735 & 1240 & 2154 & 63 & 656 & \textbf{0} & 14490 & 665 \\
\hline
\textbf{Total Answers} & 1362 & 2992 & \textbf{3000} & 2687 & 2877 & 2878 & 2945 & \textbf{3000} & 2964 & \textbf{3000} & 166 & 2888 \\
\hline
\textbf{Total Cost (USD)} & \$23.14 & \$1.44 & \$45.15 & \$40.10 & \$170.33 & \$28.79 & \$6.33 & \$1.96 & \textbf{\$0.67} & \textbf{\$0.09} & \$115.84 & \$12.75 \\
\hline
\textbf{Total Time} & 58h & 1h10m & 12h6m & 7h23m & 26h19m & 4h40m & 20h40m & 6h34m & \textbf{7h53m} & \textbf{32m29s} & 129h31m & 10h27m \\
\hline
\textbf{Efficiency} & 0.31 & 0.55 & \textbf{0.62} & 0.00 & 0.00 & 0.33 & 0.48 & 0.39 & 0.54 & \textbf{0.65} & 0.00 & 0.00 \\
\hline
\textbf{MMLU PRO (engineering)} & \multicolumn{2}{c|}{*} & \multicolumn{2}{c|}{61.53} & \multicolumn{2}{c|}{55.00} & \multicolumn{2}{c|}{39.42} & \multicolumn{2}{c|}{44.16} & \multicolumn{2}{c|}{48.71} \\
\hline
\end{tabular} 
\vspace{5pt}
\caption{Electronic Engineering Multichocie Test (EEMT) First Match Scores, Costs, Times, and Efficiency per Source and Model. \\ \footnotesize{(* MMLU PRO score was not available for this model). Results dated 2024-09-18}}
\vspace{-3mm}\end{table*}
To assess the ability of LLMs to accurately select commands for execution, a specialized test was developed using Standard Commands for Programmable Instruments (SCPI).
A total of 101 questions, for which the answer was a SCPI command, were manually created for the Keysight Oscilloscope DSOX1204G. The LLMs were evaluated based on their ability to generate the correct SCPI command. Two evaluation stages were considered: (1) when the LLM had no contextual information about the available commands for the specific model, and (2) when the LLM was provided with the full programming manual (SCPI Commands) of the instrument in its context window. Two performance metrics were employed: (1) an exact match metric, which assessed whether the LLM generated the precise SCPI command, and (2) a partial match metric, referred to as the SCPI Score, which awarded partial credit based on the closeness of the generated command to the correct SCPI command.

\section{Experimental Setup}

LLMs were given five attempts to provide the correct answer in a format readable by the evaluator. The process ran concurrently with automatic retries to manage API rate limits.

For the EEMT task, 3000 questions generated by Gemini Pro 1.5 were used. Two runs were conducted: one without Chain-of-Thought (CoT) prompting, and one with CoT prompting.

For the SCPI task, 101 questions were tested. Three runs were performed: one without CoT, one with CoT, and one using the DSOX1204G context versus no context.

\section{Results and Discussion}

The results of this study provide valuable insights into the performance of Large Language Models (LLMs) in electronic engineering tasks, particularly in the EEMT and SCPI tasks. Several key findings emerged from the experiments:

\subsection{Synthetic Benchmark and Model Quality}

It was observed that the quality of the synthetic benchmarks shown in Figure \ref{fig:family} created by LLMs is closely correlated with the overall model performance. Models generally scored higher when tested on synthetic data generated from models within their family, suggesting an inherent advantage when handling data that mirrors their own training paradigm. This observation raises questions about the robustness of models when generalizing across different domains or datasets generated by other models.

\subsection{LLMs Performance Across Tasks}

Claude Sonnet 3.5 emerged as the best-performing model in the EEMT task, demonstrating superior accuracy in ranking and selecting the most relevant responses. In contrast, GPT-4o (2024-05-13) performed best in the SCPI task, where it excelled in accurately selecting SCPI commands. However, in terms of overall efficiency—considering factors like computational cost and time—Gemini Flash provided the best value for money and operational time, making it the most efficient model across both tasks.

\subsection{Impact of Chain-of-Thought Prompting}

The inclusion of Chain-of-Thought (CoT) reasoning did not improve model performance across tasks. In fact, CoT prompting often led to a slight degradation in accuracy, suggesting that multi-step reasoning may not align well with the structured and fact-driven nature of these tasks. This could be due to \textit{cognitive overload}, where models must reason over 20 possible options, potentially causing confusion rather than clarity, thus reducing overall accuracy. As well as significantly increasing errors when attempting to extract answers, this additionally increases the cost of running the benchmark and total time taken. This process made models like Gemini 1.5 Pro extremely error prone.
Furthermore, the limited context window of 8,192 tokens restricts the model's ability to fully reason over complex inputs. This constraint could hamper detailed, domain-specific reasoning that requires sustained attention to larger inputs. These challenges are consistent with prior studies that highlight cognitive constraints in LLMs \cite{Xu2023CognitiveThinking} and performance declines due to token limitations in multi-step reasoning \cite{Wei2022Chain-of-ThoughtModels}.

\subsection{Category Performance in EEMT}

One clear trend that emerged from the EEMT task  as shown in Table 1 was the underperformance of models in instrumentation-related categories. Categories like \textit{Instrumentation}, \textit{Instrumentation Fundamentals}, and \textit{Instrumentation Principles} consistently scored lower compared to other categories such as \textit{OpAmp} and \textit{RF Circuit Design}. This suggests that LLMs currently have limited domain knowledge in the instrumentation subfield, underscoring the need for more targeted training in these specialized areas.

\subsection{LLM Knowledge of SCPI Commands}

The SCPI task  as shown in figure \ref{fig:scpi_correct} revealed that while LLMs have a limited understanding of SCPI commands in isolation, they perform remarkably well when provided with a context window that includes sparse information about all available commands and a description of their functions. 
This approach comes with a drastically higher token usage (see Figure \ref{fig:scpi_tokens}), increased total time to complete the benchmark (see Figure \ref{fig:scpi_time}) and higher LLM inference costs (see Figure \ref{fig:scpi_cost}). This ability to identify the correct command based on contextual clues demonstrates that LLMs are adept at pattern recognition and retrieval, even when their explicit domain knowledge is lacking.

\section{Limitations}

While the Test and Measurement Intelligence Quotient (TMIQ) offers a valuable framework for evaluating Large Language Models (LLMs) in the electronic engineering domain, several limitations must be considered.
 The data shows that Synthetically generated benchmark's difficultly is correlated to an LLM existing knowledge of that field. Using LLMs to generate question-answer pairs may overlook highly specialized nuances in fields like RF or VLSI. This could bias the questions toward general topics, potentially under-challenging the models. Expert review may be required to ensure question quality. LLMs, particularly within the same family, tend to perform better on synthetic benchmarks generated by themselves, as seen with Figure \ref{fig:family}. This bias could distort evaluations by favoring models that perform better on data similar to their training rather than demonstrating true domain generalization. Creating incorrect answer options involves subjective judgment, and even domain experts might disagree on their subtlety. This could introduce inconsistencies in model evaluation. The benchmark currently uses only a single set of prompts to evaluate LLM performance. Without incorporating multiple prompt variations, the benchmark may not fully capture the models' robustness to different prompt formulations. 

\section{Conclusion}
% Summary and future directions

In this paper, we introduced the Test and Measurement Intelligence Quotient (TMIQ), a specialized benchmark designed to expeditiously elevate the assessment and selection of Large Language Models (LLMs) in the Test and Measurement domain. By incorporating tasks that require precise command generation and domain-specific reasoning, TMIQ addresses the limitations of previous benchmarks in evaluating LLMs' capabilities in specialized fields. 
Our evaluations show that while LLMs demonstrate promise in selecting the correct SCPI commands, there is substantial room for improvement in their overall electronic engineering knowledge, particularly in instrumentation. The results indicate that even leading models face challenges in fully grasping the complexities of the Test and Measurement domain, highlighting the need for further research to enhance their domain-specific expertise. We hope that TMIQ will serve as a valuable tool in pushing the boundaries of what LLMs can achieve in specialized industries, ultimately contributing to the development of AI-driven solutions that meet the high standards of precision and efficiency required in electronic engineering.

\section*{Acknowledgements}

\blackout{The authors would like to thank Keysight Technologies, United States, for their funding support of this project, Royal Society grant RGS\textbackslash R2\textbackslash 222007 for their support in computing resources, EDINA and ISG@University of Edinburgh for their support in enabling access to OpenAI services.}

\newpage
% References
\small
\bibliographystyle{IEEEtran}
\bibliography{references, table}
\end{document}